\documentclass[11pt,a4paper]{article}
\usepackage[hyperref]{conll-2019}
\usepackage{times}
\usepackage{latexsym}

\usepackage{url}

\usepackage{algpseudocode}
\usepackage{algorithm}
\usepackage{graphicx}
\usepackage{amsmath}
\usepackage{dsfont}
\usepackage{color}
\usepackage{multirow}
\usepackage{multicol}
\usepackage{tabularx}
\usepackage{booktabs}
\usepackage{pbox}
\usepackage{framed}
\usepackage{array}
\usepackage{enumitem}
\usepackage{chngpage}
\usepackage{threeparttable}

\aclfinalcopy 

\newcommand{\eg}{{e.g.}}

\newcommand{\ie}{{i.e.}}

\newcommand{\hhide}[1]{}

\newcommand\blfootnote[1]{%
  \begingroup
  \renewcommand\thefootnote{}\footnote{#1}%
  \addtocounter{footnote}{-1}%
  \endgroup
}

\newcommand{\eat}[1]{\ignorespaces}

\newcommand{\ignore}[1]{}

\title{Evidence Sentence Extraction for Machine Reading Comprehension}

\author{
 Hai Wang\textsuperscript{1}\textsuperscript{*} ~~ Dian Yu\textsuperscript{2} ~~ Kai Sun\textsuperscript{3}\textsuperscript{*}  ~~ Jianshu Chen\textsuperscript{2} \\
 ~~ \textbf{Dong Yu}\textsuperscript{2} ~~ \textbf{David McAllester}\textsuperscript{1} ~~ \textbf{Dan Roth}\textsuperscript{4} \\
 \textsuperscript{1}Toyota Technological Institute at Chicago, Chicago, IL, USA \\
 \textsuperscript{2}Tencent AI Lab, Bellevue, WA, USA  \textsuperscript{3}Cornell, Ithaca, NY, USA \\
 \textsuperscript{4}University of Pennsylvania, Philadelphia, PA, USA \\
  \{haiwang,mcallester\}@ttic.edu, ks985@cornell.edu, \\
  \{yudian,jianshuchen,dyu\}@tencent.com, danroth@seas.upenn.edu  \\
}

\date{}

\begin{document}
\maketitle

\blfootnote{* This work was done when H. W. and K. S. were at Tencent AI Lab, Bellevue, WA.} 

\begin{abstract}

Remarkable success has been achieved in the last few years on some limited machine reading comprehension (MRC) tasks. However, it is still difficult to interpret the predictions of existing MRC models. In this paper, we focus on extracting evidence sentences that can explain or support the answers of multiple-choice MRC tasks, where the majority of answer options cannot be directly extracted from reference documents.

Due to the lack of ground truth evidence sentence labels in most cases, we apply distant supervision to generate imperfect labels and then use them to train an evidence sentence extractor. To denoise the noisy labels, we apply a recently proposed deep probabilistic logic learning framework to incorporate both sentence-level and cross-sentence linguistic indicators for indirect supervision. We feed the extracted evidence sentences into existing MRC models and evaluate the end-to-end performance on three challenging multiple-choice MRC datasets: MultiRC, RACE, and DREAM, achieving comparable or better performance than the same models that take as input the full reference document. To the best of our knowledge, this is the first work extracting evidence sentences for multiple-choice MRC.

\end{abstract}

\section{Introduction}
\label{sec:intro}

Recently, there have been increased interests in machine reading comprehension (MRC). In this work, we mainly focus on multiple-choice MRC~\cite{richardson2013mctest,mostafazadeh2016corpus,ostermann2018semeval}: given a document and a question, the task aims to select the correct answer option(s) from a small number of answer options associated with this question. Compared to extractive and abstractive MRC tasks (\eg,~\cite{rajpurkar2016squad,kovcisky2018narrativeqa,reddy2019coqa}) where most questions can be answered using spans from the reference documents, the majority of answer options cannot be directly extracted from the given texts.

Existing multiple-choice MRC models~\cite{wang2018co,radfordimproving} take as input the entire reference document and seldom offer any explanation, making interpreting their predictions extremely difficult. It is a natural choice for human readers to use sentences from a given text to explain why they select a certain answer option in reading tests~\cite{bax2013cognitive}. In this paper, as a preliminary attempt, we focus on exacting \emph{\textbf{evidence sentences}} that entail or support a question-answer pair from the given reference document.

For extractive MRC tasks, information retrieval techniques can be very strong baselines to extract sentences that contain questions and their answers when questions provide sufficient information, and most questions are factoid and answerable from the content of a single sentence~\cite{lin2018denoising,min2018efficient}. However, we face unique challenges to extract evidence sentences for multiple-choice MRC tasks. The correct answer options of a significant number of questions (\eg, $87\%$ questions in RACE~\cite{lai2017race,sundream2018}) are not extractive, which may require advanced reading skills such as inference over multiple sentences and utilization of prior knowledge~\cite{lai2017race,khashabi2018looking,ostermann2018semeval}. Besides, the existence of misleading wrong answer options also dramatically increases the difficulty of evidence sentence extraction, especially when a question provides insufficient information. For example, in Figure~\ref{fig:overview}, given the reference document and question \emph{``Which of the following statements is true according to the passage?''}, almost all the tokens in the wrong answer option B \emph{``In 1782, Harvard began to teach German.''} appear in the document (\ie, sentence S$_9$ and S$_{11}$) while the question gives little useful information for locating answers. Furthermore, we notice that even humans sometimes have difficulty in finding pieces of evidence when the relationship between a question and its correct answer option is implicitly indicated in the document (\eg, \emph{``What is the main idea of this passage?''}). Considering these challenges, we argue that extracting evidence sentences for multiple-choice MRC is at least as difficult as that for extractive MRC or factoid question answering.

Given a question, its associated answer options, and a reference document, we propose a method to extract sentences that can potentially support or explain the (question, correct answer option) pair from the reference document. Due to the lack of ground truth evidence sentences in most multiple-choice MRC tasks, inspired by distant supervision, we first extract \emph{silver standard} evidence sentences based on the lexical features of a question and its correct answer option (Section~\ref{sec:noisy_gt}), then we use these noisy labels to train an evidence sentence extractor (Section~\ref{sec:transformer}). To denoise imperfect labels, we also manually design sentence-level and cross-sentence linguistic indicators such as \emph{``adjacent sentences tend to have the same label''} and accommodate all the linguistic indicators with a recently proposed deep probabilistic logic learning framework~\cite{haidpl2018} for indirect supervision (Section~\ref{sec:dpl}). 

Previous extractive MRC and question answering studies~\cite{min2018efficient,lin2018denoising} indicate that a model should be able to achieve comparable end-to-end performance if it can accurately predict the evidence sentence(s). Inspired by the observation, to indirectly evaluate the quality of the extracted evidence sentences, we only keep the selected sentences as the new reference document for each instance and evaluate the performance of a machine reader~\cite{wang2018co,radfordimproving} on three challenging multiple-choice MRC datasets: MultiRC~\cite{khashabi2018looking}, RACE~\cite{lai2017race}, and DREAM~\cite{sundream2018}. Experimental results show that we can achieve comparable or better performance than the same reader that considers the full context. The comparison between ground truth evidence sentences and automatically selected sentences indicates that there is still room for improvement.

Our primary contributions are as follows: 1) to the best of our knowledge, this is the first work to extract evidence sentences for multiple-choice MRC; 2) we show that it may be a promising direction to leverage various sources of linguistic knowledge for denoising noisy evidence sentence labels. We hope our attempts and observations can encourage the research community to develop more explainable MRC models that simultaneously provide predictions and textual evidence.

\section{Method}

\begin{figure}[h!]
   \begin{center}
   \includegraphics[width=0.5\textwidth]{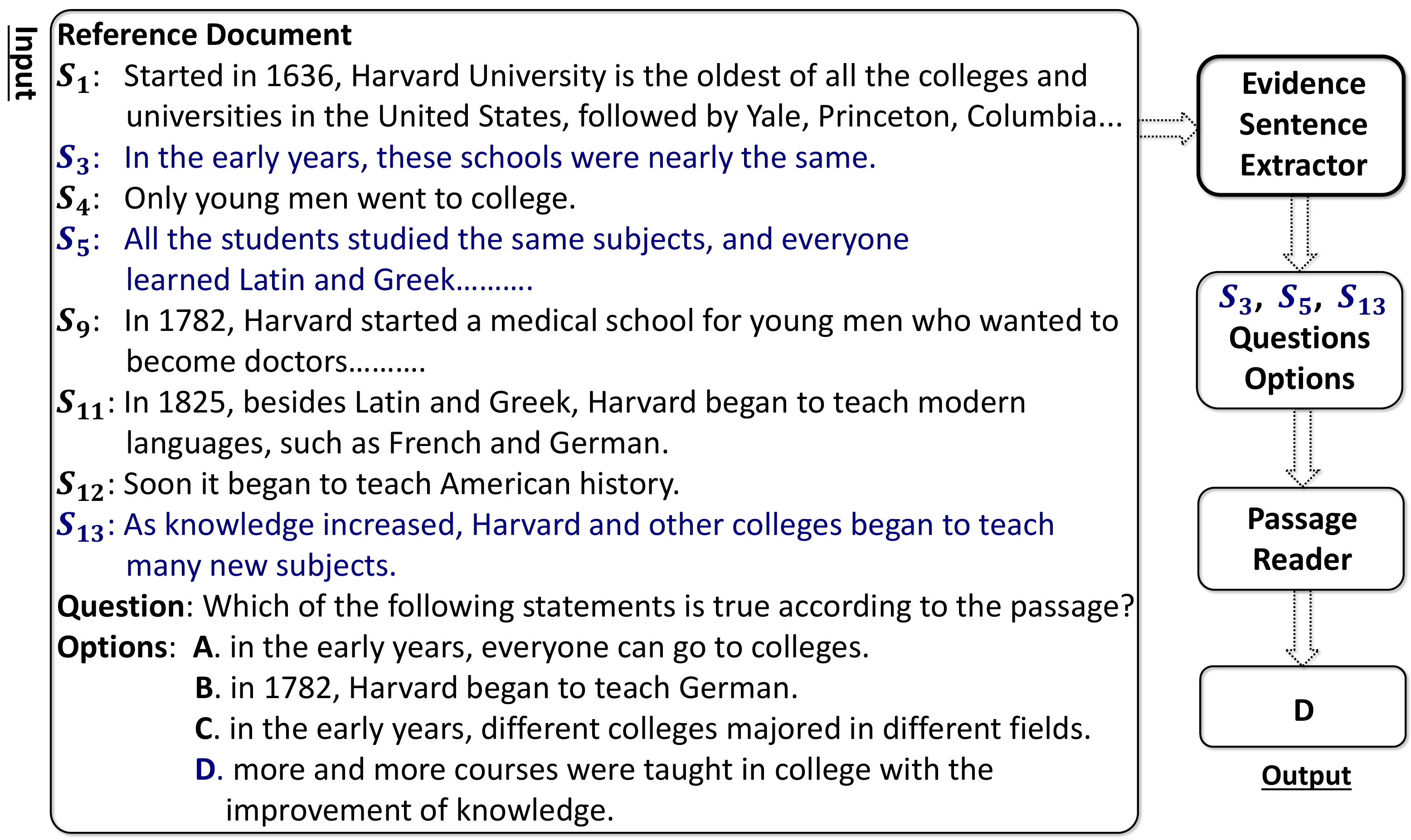}
   \end{center}
 \caption{An overview of our pipeline. The input instance comes from RACE~\cite{lai2017race}.}
 \label{fig:overview}
\end{figure}


We will present our evidence sentence extractor (Section~\ref{sec:transformer}) trained on the noisy training data generated by distant supervision (Section~\ref{sec:noisy_gt}) and denoised by an existing deep probabilistic logic framework that incorporates different kinds of linguistic indicators (Section~\ref{sec:dpl}). The extractor is followed by an independent neural reader for evaluation. See an overview in Figure~\ref{fig:overview}.


\subsection{Evidence Sentence Extractor}
\label{sec:transformer}
We use a multi-layer multi-head transformer~\cite{vaswani2017attention} to extract evidence sentences.
Let $W_{w}$ and $W_{p}$ be the word (subword) and position embeddings, respectively. Let $M$ denote the total number of layers in the transformer. Then, the $m$-th layer hidden state $h^{m}$ of a token is given by: 

\begin{equation}
\small
h^m = \begin{cases}
W_{w} + W_{p} &\text{if $m=0$}\\
\text{TB}(h^{m-1})  &\text{if $1\leq m \leq M$}
\end{cases}
\end{equation}
where TB stands for the Transformer Block, which is a standard module that contains MLP, residual connections~\cite{he2016deep} and LayerNorm~\cite{ba2016layer}.

Compared to classical transformers, pre-trained transformers such as GPT~\cite{radfordimproving} and BERT~\cite{devlin2018bert} capture rich world and linguistic knowledge from large-scale external corpora, and significant improvements are obtained by fine-tuning these pre-trained models on a variety of downstream tasks. We follow this promising direction by fine-tuning GPT~\cite{radfordimproving} on a target task. Note that the pre-trained transformer in our pipeline can also be easily replaced with other pre-trained models, which however is not the focus of this paper.

We use $(X, Y)$ to denote all training data, $(X_{i}, Y_{i})$ to denote each instance, where $X_{i}$ is a token sequence, namely, $X_{i} = \{X_{i}^{1},\ldots, X_{i}^{t} \}$ where $t$ equals to the sequence length. For evidence sentence extraction, $X_{i}$ contains one sentence in a document, a question, and all answer options associated with the question. $Y_{i}$ indicates the probability that sentence in $X_{i}$ is selected as an evidence sentence for this question, and $\sum_{i=1}^{N}Y_{i}=1$, where $N$ equals to the total number of sentences in a document. GPT takes as input $X_{i}$ and produces the final hidden state $h_{i}^{M}$ of the last token in $X_{i}$, which is further fed into a linear layer followed by a softmax layer to generate the probability:
\begin{eqnarray}
 P_{i} = \frac{\text{exp}(W_{y}h_{i}^{M})}{\sum_{1 \leq i \leq N}\text{exp}(W_{y}h_{i}^{M})}
\end{eqnarray}
where $W_{y}$ is weight matrix of the output layer. We use Kullback-Leibler divergence loss $\text{KL}(Y||P)$ as the training criterion.

\hhide{
We use $(X, Y)$ to denote all training data, $(X_{i}, Y_{i})$ to denote each instance, where $X_{i}$ is a token sequence, namely, $X_{i} = \{X_{i}^{1},\ldots, X_{i}^{t} \}$ where $t$ equals to the sequence length. For evidence sentence extraction, $X_{i}$ contains one sentence in a document, a question, and all answer options associated with the question. $Y_{i}$ indicates the probability that sentence $X_{i}$ is selected as an evidence sentence for this question, and $\sum_{i=1}^{N}Y_{i}=1$, where $N$ equals to the total number of sentences in a document. GPT takes as input $X_{i}$ and produces the final hidden state $h_{i}^{M}$ of the last token in $X_{i}$, which is further fed into a linear layer followed by a softmax layer to generate the probability:
\begin{eqnarray}
 P_{i} = \frac{\text{exp}(W_{y}h_{i}^{M})}{\sum_{1 \leq i \leq N}\text{exp}(W_{y}h_{i}^{M})}
\end{eqnarray}
where $W_{y}$ is weight matrix of the output layer. We use Kullback-Leibler divergence loss $\text{KL}(Y||P)$ as the training criterion.
}


We first apply distant supervision to generate noisy evidence sentence labels (Section~\ref{sec:noisy_gt}). To denoise the labels, during the training stage, we use deep probabilistic logic learning (DPL) -- a general framework for combining indirect supervision strategies by composing probabilistic logic with deep learning~\cite{haidpl2018}. Here we consider both sentence-level and cross-sentence linguistic indicators as indirect supervision strategies (Section~\ref{sec:dpl}).

\begin{figure*}[!ht]
   \begin{center}
   \includegraphics[width=0.93\textwidth]{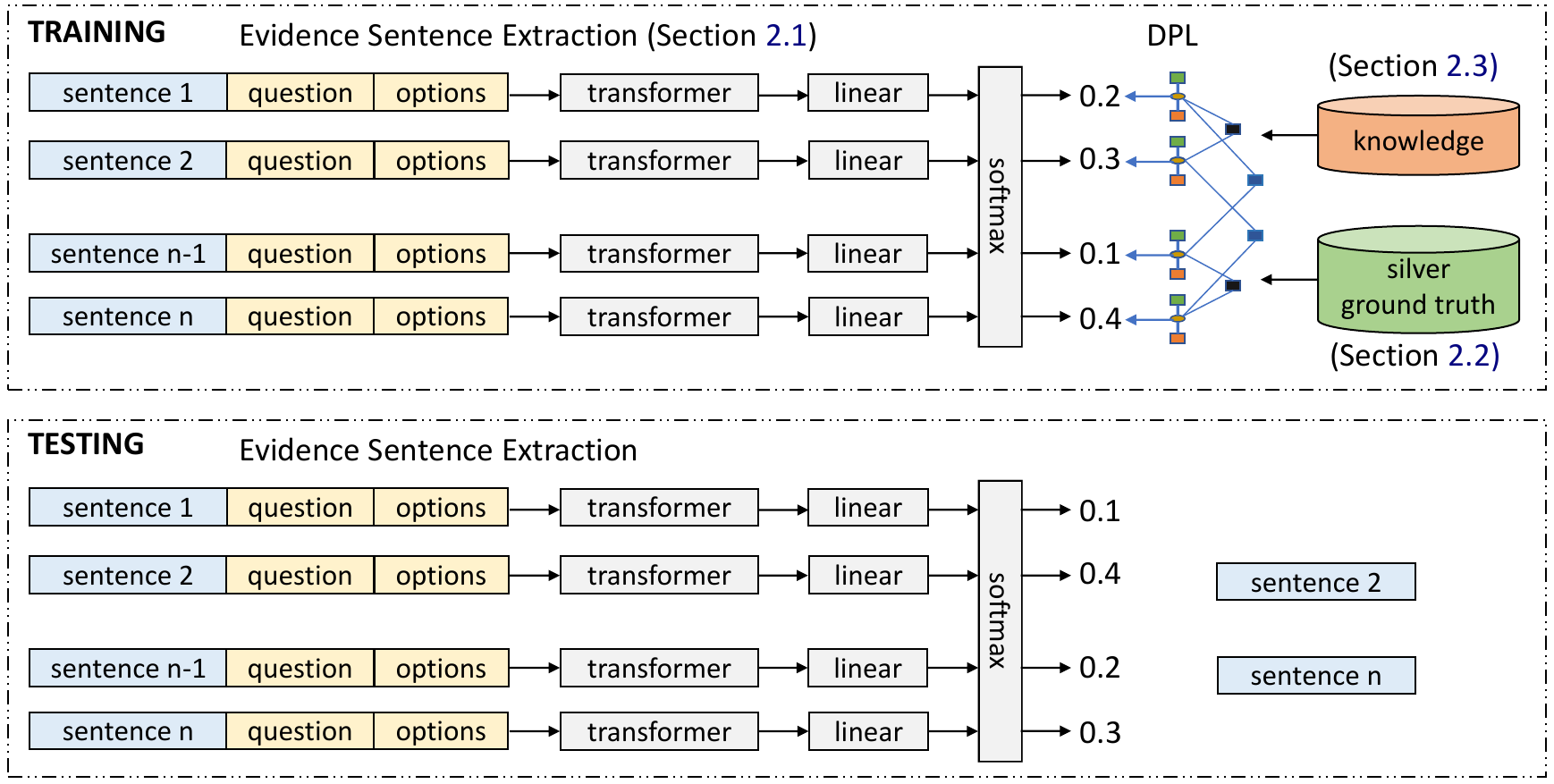}
   \end{center}
 \caption{Deep probabilistic logic (DPL) framework for evidence sentence extraction. During testing, we only use trained evidence sentence extractor for prediction.}
 \label{fig:method:nnStruct}
\end{figure*}

As shown in Figure \ref{fig:method:nnStruct}, during training, our evidence sentence extractor contains two components: a probabilistic graph containing various sources of indirect supervision used as a supervision module and a fine-tuned GPT used as a prediction module. The two components are connected via a set of latent variables indicating whether each sentence is an evidence sentence or not. We update the model by alternatively optimizing GPT and the probabilistic graph so that they reach an agreement on latent variables. After training, only the fine-tuned GPT is kept to make predictions for a new instance during testing. We provide more details in Appendix~\ref{sec:appendix} and refer readers to~\newcite{haidpl2018} for how to apply DPL as a tool in a downstream task such as relation extraction.






\subsection{Silver Standard Evidence Generation}
\label{sec:noisy_gt}

Given correct answer options, we use a distant supervision method to generate the \emph{silver standard} evidence sentences.  

Inspired by Integer Linear Programming models (ILP) for summarization~\cite{berg2011jointly,boudin2015concept}, we model evidence sentence extraction as a maximum coverage problem and define the value of a selected sentence set as the sum of the weights for the unique words it contains. Formally, let $v_i$ denote the weight of word $i$, $v_i=1$ if word $i$ appears in the correct answer option,  $v_i=0.1$ if it appears in the question but not in the correct answer option, and $v_i=0$ otherwise.\footnote{We do not observe a significant improvement by tuning parameters $v_i$ on the development set.}

We use binary variables $c_i$ and $s_j$ to indicate the presence of word $i$ and sentence $j$ in the selected sentence set, respectively. $\text{Occ}_{i,j}$ is a binary variable indicating the occurrence of word $i$ in sentence $j$, $l_j$ denotes the length of sentence $j$, and $L$ is the predefined maximum number of selected sentences. We formulate the ILP problem as: 
\begin{eqnarray}
\small
 \max \sum_i v_i c_i \quad  s.t. \sum_j s_j \leq L \\
 s_j ~\text{Occ}_{ij} \leq c_i,  \forall i,j \quad \sum_{j} s_j ~\text{Occ}_{ij} \geq c_i, \forall i \\ \nonumber
c_i \in \{0, 1\} \ \forall i \quad s_j \in \{0, 1\} \ \forall j \nonumber
\end{eqnarray}

\subsection{Linguistic Indicators for Indirect Supervision}
\label{sec:dpl}

To denoise the imperfect labels generated by distant supervision (Section~\ref{sec:noisy_gt}), as a preliminary attempt, we manually design a small number of sentence-level and cross-sentence linguistic indicators incorporated in DPL for indirect supervision. We briefly introduce them as follows and detail all indicators in Appendix~\ref{sec:strategies} and implementation details in Section~\ref{sec:implementation}.

We assume that a sentence is more likely to be an evidence sentence if the sentence and the question have similar meanings, lengths, coherent entity types, same sentiment polarity, or the sentence is true (\ie, entailment) given the question. We assume that a good evidence sentence should be neither too long nor too short (\ie, $5 \leq \#~\text{of tokens in sentence} \leq 40$) considering informativeness and conciseness, and an evidence sentence is more likely to lead to the prediction of the correct answer option (referred as ``reward''), which is motivated by our experiments that machine readers take as input the silver (or gold) standard evidence sentences achieve the best performance except for human performance on three multiple-choice machine reading comprehension datasets (Table~\ref{tab:multirc_result}, Table~\ref{tab:race_result}, and Table~\ref{tab:dream_result} in Section~\ref{sec:exp}). We rely on both lexical features (\eg, lengths and entity types) and semantic features based on pre-trained word/paraphrase embeddings and external knowledge graphs to measure the similarity of meanings. We use existing models or resources for reward calculation, sentiment analysis and natural language inference.

For cross-sentence indicators, we consider that the same set of evidence sentences are less likely to support multiple questions and two evidence sentences that support the same question should be within a certain distance (\ie, evidence sentences for the same question should be within window size $8$ (in sentences)). We also assume that adjacent sentences tend to have the same label. We will have more discussions about these assumptions in the data analysis (Section~\ref{sec:human_eval}).

\section{Experiments}
\label{sec:exp}

\begin{table*}[ht!]
\footnotesize
\centering
\begin{tabular}{lccccccc}
\toprule
\multirow{2}{*}{\textbf{Dataset}} & \multicolumn{3}{c}{\textbf{\# of documents}} & \multicolumn{3}{c}{\textbf{\# of questions}} & \textbf{Average \# of sentences per document} \\
                         & Train& Dev & Test& Train & Dev & Test & Train + Dev + Test \\
\midrule
MultiRC                  &  456       & 83         & 332     & 5,131       &  953  & 3,788 & 14.5 (Train + Dev)    \\
DREAM                    & 3,869      & 1,288     & 1,287     & 6,116       & 2,040        & 2,041 & 8.5\\
RACE                     & 25,137     & 1,389     & 1,407     & 87,866      & 4,887        & 4,934  & 17.6   \\
\bottomrule
\end{tabular}
\caption{Statistics of multiple-choice machine reading comprehension and question answering datasets.}
\label{tab:dataset_statistics}
\end{table*}

\subsection{Datasets}

We use the following three latest multiple-choice machine reading comprehension datasets for evaluation. We show data statistics in Table~\ref{tab:dataset_statistics}.


\noindent \textbf{MultiRC}~\cite{khashabi2018looking}: MultiRC is a dataset in which questions can only be answered by considering information from multiple sentences. A question may have multiple correct answer options. Reference documents come from seven different domains such as elementary school science and travel guides. For each document, questions and their associated answer options are generated and verified by turkers.

\noindent \textbf{RACE}~\cite{lai2017race}: RACE is a dataset collected from English language exams designed for middle (RACE-Middle) and high school (RACE-High) students in China, carefully designed by English instructors. The proportion of questions that requires reasoning is $59.2\%$.

\noindent \textbf{DREAM}~\cite{sundream2018}: DREAM is a dataset collected from English exams for Chinese language learners. Each instance in DREAM contains a multi-turn multi-party dialogue, and the correct answer option must be inferred from the dialogue context. In particular, a large portion of questions require multi-sentence inference ($84\%$) and/or commonsense knowledge ($34\%$). 

\subsection{Implementation Details}
\label{sec:implementation}

We use spaCy~\cite{honnibal2015improved} for tokenization and named entity tagging. We use the pre-trained transformer (\ie, GPT) released by~\newcite{radfordimproving} with the same pre-processing procedure. When GPT is used as the neural reader, we set training epochs to $4$, use eight P40 GPUs for experiments on RACE, and use one GPU for experiments on other datasets. When GPT is used as the evidence sentence extractor, we set batch size $1$ per GPU and dropout rate $0.3$. We keep other parameters default. Depending on the dataset, training the evidence sentence extractor generally takes several hours. 


For DPL, we adopt the toolkit from~\newcite{haidpl2018}. During training, we conduct message passing in DPL iteratively, which usually converges within $5$ iterations. We use Vader~\cite{gilbert2014vader} for sentiment analysis and ParaNMT-$50$M~\cite{wieting2018paranmt} to calculate the paraphrase similarity between two sentences. We use the knowledge graphs (\ie, triples in ConceptNet v$5.6$~\cite{speer2012representing,speer2017conceptnet}) to incorporate commonsense knowledge. To calculate the natural language inference probability, we first fine-tune the transformer~\cite{radfordimproving} on several tasks, including SNLI~\cite{bowman2015large}, SciTail~\cite{khot2018scitail}, MultiNLI~\cite{williams2018broad}, and QNLI~\cite{wang2018glue}.

To calculate the probability that each sentence leads to the correct answer option, we sample a subset of sentences and use them to replace the full context in each instance, and then we feed them into the transformer fine-tuned with instances with full context. If a particular combination of sentences $S=\{s_{1},\ldots,s_{n}\}$ leads to the prediction of the correct answer option, we reward each sentence inside this set with $1/n$. To avoid the combinatorial explosion, we assume evidence sentences lie within window size $3$. For another neural reader Co-Matching~\cite{wang2018co}, we use its default parameters. For DREAM and RACE, we set $L$, the maximum number of silver standard evidence sentences of a question, to $3$. For MultiRC, we set $L$ to 5 since many questions have more than $5$ ground truth evidence sentences.

\begin{table*}[!htp]
\centering
\footnotesize
    \begin{tabular}{lccc}
    \toprule
     \textbf{Approach} & $\text{F1}_{m}$ & $\text{F1}_{a}$ & $\text{EM}_{0}$\\ 
     \midrule
     All-ones baseline \cite{khashabi2018looking} & 61.0 & 59.9 & 0.8 \\
     Lucene world baseline \cite{khashabi2018looking} & 61.8 & 59.2 & 1.4 \\
     Lucene paragraphs baseline \cite{khashabi2018looking} & 64.3 &60.0 &7.5\\
     Logistic regression \cite{khashabi2018looking} & 66.5	& 63.2 & 11.8\\
     Full context + Fine-Tuned Transformer (GPT,~\newcite{radfordimproving}) & 68.7	& 66.7 & 11.0\\
     \midrule
     Random 5 sentences + GPT  & 65.3	& 63.1 & 7.2	\\
     Top 5 sentences by $\text{ESE}_{\text{DS}}$~$\text{+ GPT}$  & 70.2 & \textbf{68.6} & 12.7 \\
     Top 5  sentences by $\text{ESE}_{\text{DPL}}$~$\text{+ GPT}$ & \bf 70.5 & 67.8 & \bf 13.3 \\
     \midrule
     Top 5 sentences by $\text{ESE}_{\text{gt}}$~$\text{+ GPT}$ & \textbf{72.3} & \textbf{70.1} & \textbf{19.2} \\
     \midrule
     Ground truth evidence sentences + GPT  & 78.1 & 74.0 & 28.6 \\
     Human Performance \cite{khashabi2018looking} & 86.4  & 83.8  &56.6 \\
     \bottomrule
  \end{tabular}
  \caption{Performance of various settings on the MultiRC development set. We use the fine-tuned GPT as the evidence sentence extractor (ESE) and the neural reader ($\text{ESE}_{\text{DS}}$: ESE trained on the silver standard evidence sentences;
  $\text{ESE}_{\text{DPL}}$: ESE trained with DPL as a supervision module; $\text{ESE}_{\text{gt}}$: ESE trained using ground truth evidence sentences; $\text{F1}_\text{{m}}$\: macro-average F1; $\text{F1}_\text{{a}}$: micro-average F1; $\text{EM}_\text{{0}}$: exact match).
  }
  \label{tab:multirc_result}
 \end{table*}

\subsection{Evaluation on MultiRC} 
\label{eval:multirc}

Since its test set is not publicly available, currently we only evaluate our model on the development set (Table~\ref{tab:multirc_result}). The fine-tuned transformer (GPT) baseline, which takes as input the full document, achieves an improvement of $2.2\%$ in macro-average F1 ($\text{F1}_{\text{m}}$) over the previous highest score, $66.5\%$. If we train our evidence sentence extractor using the ground truth evidence sentences provided by turkers, we can obtain a much higher $\text{F1}_{\text{m}}$ $72.3\%$, even after we remove nearly $66\%$ of sentences in average per document. We can regard this result as the supervised upper bound for our evidence sentence extractor. If we train the evidence sentence extractor with DPL as a supervision module, we get $70.5\%$ in $\text{F1}_{\text{m}}$. The performance gap between $70.5\%$ and $72.3\%$ shows there is still room for improving denoising strategies.



\begin{center}
\begin{table*}[!h]
\centering
\footnotesize
\begin{tabular}{lcccccc}
\toprule
     \multirow{2}{*}{\textbf{Approach}} & \multicolumn{3}{c}{\textbf{Dev}} & \multicolumn{3}{c}{\textbf{Test}}\\ 
     & Middle & High & All & Middle & High & All \\ 
     \midrule
     Sliding Window~\cite{richardson2013mctest,lai2017race} & - & - &-  & 37.3 & 30.4 & 32.2 \\ 
     Co-Matching~\cite{wang2018co} & - & - & - & 55.8 & 48.2 & 50.4 \\
     Full context + GPT~\cite{radfordimproving} & - & - & - & 62.9 & 57.4 & 59.0 \\
     \midrule
     Full context + GPT  & 55.6 & 56.5 & 56.0 & 57.5 & 56.5 & 56.8 \\
     Random 3 sentences + GPT & 50.3 & 51.1 & 50.9  & 50.9 & 49.5 & 49.9 \\ 
     \midrule
     Top 3 sentences by InferSent (question) + Co-Matching & 49.8 & 48.1 & 48.5 & 50.0 & 45.5 & 46.8 \\
     Top 3 sentences by InferSent (question + all options) + Co-Matching & 52.6 & 49.2 & 50.1  & 52.6 & 46.8 & 48.5 \\ 
     Top 3 sentences by $\text{ESE}_{\text{DS}}$ + Co-Matching & 58.1 & 51.6 & 53.5 & 55.6 & 48.2 & 50.3 \\
     Top 3 sentences by $\text{ESE}_{\text{DPL}}$ + Co-Matching & 57.5 & 52.9 & 54.2  & 57.5 & 49.3 & 51.6 \\
     \midrule
     Top 3 sentences by InferSent (question) + GPT & 55.0 & 54.7 & 54.8  & 54.6 & 53.4 &   53.7 \\ 
     Top 3 sentences by InferSent (question + all options) + GPT & 59.2 & 54.6 & 55.9  & 57.2 & 53.8 & 54.8   \\ 
     Top 3 sentences by $\text{ESE}_{\text{DS}}$ + GPT & 62.5 & 57.7 &  59.1 & 64.1 & 55.4 & 58.0 \\ 
     Top 3 sentences by $\text{ESE}_{\text{DPL}}$ + GPT & 63.2 & 56.9 & 58.8 & \textbf{64.3} & 56.7 & 58.9 \\
     \midrule
     Top 3 sentences by $\text{ESE}_{\text{DS}}$ + full context + GPT & 63.4 & 58.6 & 60.0 & 63.7 & 57.7 & 59.5 \\
     Top 3 sentences by $\text{ESE}_{\text{DPL}}$ + full context + GPT & 64.2 & 58.5 & 60.2 & 62.4 & \textbf{58.7} & \textbf{59.8} \\ 
     \midrule
     Silver standard evidence sentences + GPT & 73.2 & 73.9 &  73.7 & 74.1  & 72.3 & 72.8  \\ 
     Amazon Turker Performance~\cite{lai2017race} & - & - & - & 85.1  & 69.4 & 73.3  \\ 
     Ceiling Performance~\cite{lai2017race} & - & - & - & 95.4 & 94.2 & 94.5  \\ 
     \bottomrule
  \end{tabular}
  \caption{Accuracy (\%) of various settings on the RACE dataset. $\text{ESE}_{\text{DS}}$: evidence sentence extractor trained on the silver standard evidence sentences extracted from the rule-based distant supervision method.}
  \label{tab:race_result}
 \end{table*}
\end{center}

\subsection{Evaluation on RACE}
\label{sec:race}

As we cannot find any public implementations of recently published independent sentence selectors, we compare our evidence sentence extractor with InferSent released by~\newcite{conneau-EtAl} as previous work~\cite{htut2018training} has shown that it outperforms many state-of-the-art sophisticated sentence selectors on a range of tasks. We also investigate the \emph{\textbf{portability}} of our evidence sentence extractor by combing it with two neural readers. Besides the fine-tuned GPT baseline, we use Co-Matching~\cite{wang2018co}, another state-of-the-art neural reader on the RACE dataset.

As shown in Table~\ref{tab:race_result}, by using the evidence sentences selected by InferSent, we suffer up to a $1.9\%$ drop in accuracy with Co-Matching and up to a $4.2\%$ drop with the fine-tuned GPT. In comparison, by using the sentences extracted by our sentence extractor, which is trained with DPL as a supervision module, we observe a much smaller decrease ($0.1\%$) in accuracy with the fine-tuned GPT baseline, and we slightly improve the accuracy with the Co-Matching baseline. For questions in RACE, introducing the content of answer options as additional information for evidence sentence extraction can narrow the accuracy gap, which might be due to the fact that many questions are less informative~\cite{Yichong2018dynamic}. Note that all these results are compared with $59\%$ reported from~\newcite{radfordimproving}, if compared with our own replication ($56.8\%$), sentence extractor trained with either DPL or distant supervision leads to gain up to $2.1\%$.

Since the problems in RACE are designed for human participants that require advanced reading comprehension skills such as the utilization of external world knowledge and in-depth reasoning, even human annotators sometimes have difficulties in locating evidence sentences (Section~\ref{sec:human_eval}). Therefore, \emph{\textbf{a limited number of evidence sentences might be insufficient for answering challenging questions}}. Instead of removing ``non-relevant'' sentences, we keep all the sentences in a document while adding a special token before and after the extracted evidence sentences. With DPL as a supervision module, we see an improvement in accuracy of $0.9\%$ (from $58.9\%$ to $59.8\%$). 

For our current supervised upper bound (\ie, assuming we know the correct answer option, we find the silver evidence sentences from rule-based distant supervision and then feed them into the fine-tuned transformer, we get $72.8\%$ in accuracy, which is quite close to the performance of Amazon Turkers. However, it is still much lower than the ceiling performance. To answer questions that require external knowledge, \emph{\textbf{it might be a promising direction to retrieve evidence sentences from external resources}}, compared to only considering sentences within a reference document for multiple-choice machine reading comprehension tasks.

\subsection{Evaluation on DREAM}


See Table~\ref{tab:dream_result} for results on the DREAM dataset. The fine-tuned GPT baseline, which taks as input the full document, achieves $55.1\%$ in accuracy on the test set. If we train our evidence sentence extractor with DPL as a supervision module and feed the extracted evidence sentences to the fine-tuned GPT, we get test accuracy $57.7\%$. Similarly, if we train the evidence sentence extractor only with silver standard evidence sentences extracted from the rule-based distant supervision method, we obtain test accuracy $56.3\%$, \ie, $1.4\%$ lower than that with full supervision. Experiments demonstrate the effectiveness of our evidence sentence extractor with denoising strategy, and the usefulness of evidence sentences for dialogue-based machine reading comprehension tasks in which reference documents are less formal compared to those in RACE and MultiRC.

\begin{table}[h!]
\footnotesize
\centering
  \begin{tabular}{lcc}
    \toprule
    \textbf{Approach} & \textbf{Dev} & \textbf{Test} \\ 
    \midrule
    Full context~\text{+ GPT}$^{\dag}$~\cite{sundream2018} & 55.9 & 55.5 \\
    \midrule
    Full context~\text{+ GPT}                     & 55.1 & 55.1 \\
    Top 3 sentences by $\text{ESE}_{\text{silver-gt}}$~$\text{+ GPT}$ & 50.1 & 50.4 \\
    Top 3 sentences by $\text{ESE}_{\text{DS}}$~$\text{+ GPT}$ & 55.1 & 56.3 \\
    Top 3 sentences by $\text{ESE}_{\text{DPL}}$~$\text{+ GPT}$ & 57.3 & \textbf{57.7} \\
    \midrule
    Silver standard evidence sentences~$\text{+ GPT}$ & 60.5 & 59.8 \\
    Human Performance$^{\dag}$ & 93.9 & 95.5 \\
    \bottomrule
  \end{tabular}
\caption{Performance in accuracy (\%) on the DREAM dataset (Results marked with $^{\dag}$ are taken from~\newcite{sundream2018}; $\text{ESE}_{\text{silver-gt}}$: ESE trained using silver standard evidence sentences).}
\label{tab:dream_result}
\end{table}

\subsection{Human Evaluation}
\label{sec:human_eval}

Extracted evidence sentences, which help neural readers to find correct answers, may still fail to convince human readers. Thus we evaluate the quality of extracted evidence sentences based on human annotations (Table~\ref{tab:human_eval_result}).

\begin{table}[h!]
\centering
\footnotesize
\begin{tabular}{lccc}
\toprule
\textbf{Dataset} & Silver Sentences & Sentences by $\text{ESE}_{\text{DPL}}$  \\ 
\midrule
RACE-M & 59.9  & 57.5  \\ 
MultiRC & 53.0 & 60.8 \\
\bottomrule
\end{tabular}
\caption{Macro-average F1 compared with human annotated evidence sentences on the dev set (silver sentences: evidence sentences extracted by ILP (Section~\ref{sec:noisy_gt}); sentences by $\text{ESE}_{\text{DPL}}$: evidence sentences extracted by ESE trained on silver stand ground truth, GT: ground truth evidence sentences).}
\label{tab:human_eval_result}
\end{table}

\noindent \textbf{MultiRC}: Even trained using the noisy labels, we achieve a macro-average F1 score $60.8\%$ on MultiRC, indicating the learning and generalization capabilities of our evidence sentence extractor, compared to $53.0\%$, achieved by using the noisy silver standard evidence sentences guided by correct answer options. 



\noindent \textbf{RACE}: Since RACE does not provide the ground truth evidence sentences, to get the ground truth evidence sentences, two human annotators annotate $500$ questions from the RACE-Middle development set.\footnote{Annotations are available at \url{https://github.com/nlpdata/evidence}.} The Cohen's kappa coefficient between two annotations is $0.87$. For negation questions which include negation words (\eg, \emph{``Which statement is not true according to the passage?''}), we have two annotation strategies: we can either find sentences that can directly imply the correct answer option; or the sentences that support the wrong answer options. During annotation, for each question, we use the strategy that leads to fewer evidence sentences.

\emph{\textbf{We find that even humans have troubles in locating evidence sentences when the relationship between a question and its correct answer option is implicitly implied}}. For example, a significant number of questions require the understanding of the entire document (\eg, \emph{``what's the best title of this passage''} and \emph{``this passage mainly tells us that \_''}) and/or external knowledge (\eg, \emph{``the writer begins with the four questions in order to \_''}, \emph{``The passage is probably from \_''} , and \emph{``If the writer continues the article, he would most likely write about\_''}). For $10.8\%$ of total questions, at least one annotator leave the slot blank due to the challenges mentioned above. $65.2\%$ of total questions contain at least two evidence sentences, and $70.9\%$ of these questions contain at least one adjacent sentence pair in their evidence sentences, which may provide evidence to support our assumption \emph{adjacent sentences tend to have the same label} in Section~\ref{sec:dpl}.

The average and the maximum number of evidence sentences for the remaining questions is $2.1$ and $8$, respectively. The average number of evidence sentences in the full RACE dataset should be higher since questions in RACE-High are more difficult~\cite{lai2017race}, and we ignore $10.8\%$ of the total questions that require the understanding of the whole context.

\subsection{Error Analysis}
\label{sec:error_analysis}
 
We analyze the predicted evidence sentences for instances in RACE for error analysis. Tough with a high macro-average recall ($67.9\%$), it is likely that our method extracts sentences that support distractors. For example, to answer the question \emph{``You lost your keys. You may call \_"}, our system mistakenly extracts sentences \emph{``Please call 5016666''} that support one of the distractors and adjacent to the correct evidence sentences \emph{``Found a set of keys. Please call Jane at 5019999.''} in the given document. We may need linguistic constraints or indicators to filter out irrelevant selected sentences instead of simply setting a hard length constraint such as $5$ for all instances in a dataset. 

Besides, it is possible that there is no clear sentence in the document for justifying the correctness of the correct answer. For example, to answer the question \emph{``What does ``figure out" mean ?''}, neither \emph{``find out''} nor the correct answer option appears in the given document as this question mainly assesses the vocabulary acquisition of human readers. Therefore, all the extracted sentences (\eg,  \emph{``sometimes... sometimes I feel lonely, like I'm by myself with no one here."}, \emph{``sometimes I feel excited, like I have some news I have to share!"}) by our methods are inappropriate. A possible solution is to predict whether a question is answerable following previous work (\eg,~\cite{hu2019read+}) on addressing unanswerable questions in extractive machine reading comprehension tasks such as SQuAD~\cite{rajpurkar2018squad} before to extract the evidence sentences for this question.

\section{Related Work}

\subsection{Sentence Selection for Machine Reading Comprehension and Fact Verification}

Previous studies investigate paragraph retrieval for factoid question answering~\cite{chen2017reading,wang2017r,choi2017coarse,lin2018denoising}, sentence selection for machine reading comprehension~\cite{hewlett2017accurate,min2018efficient}, and fact verification~\cite{yin2018twowingos,hanselowski2018ukp}. In these tasks, most of the factual questions/claims provide sufficient clues for identifying relevant sentences, thus often information retrieval combined with filters can serve as a very strong baseline. For example, in the FEVER dataset~\cite{thorne2018fever}, only $16.8\%$ of claims require composition of multiple evidence sentences. For some of the cloze-style machine reading comprehension tasks such as CBT~\cite{hill2015goldilocks},~\newcite{kaushik2018much} demonstrate that for some models, comparable performance can be achieved by considering only the last sentence that usually contains the answer. Different from above work, we exploit information in answer options and use various indirect supervision to train our evidence sentence extractor, and previous work can actually be a regarded as a special case for our pipeline. Compared to~\newcite{lin2018denoising}, we leverage rich linguistic knowledge for denoising imperfect labels. 

Several work also investigate content selection at the token level~\cite{yu2017learning,seo2017neural}, in which some tokens are automatically skipped by neural models. However, they do not utilize any linguistic knowledge, and a set of discontinuous tokens has limited explanation capability.

\subsection{Machine Reading Comprehension with External Linguistic Knowledge}

Linguistic knowledge such as coreference resolution, frame semantics, and discourse relations is widely used to improve machine comprehension~\cite{wang2015machine,sachan2015learning,narasimhan2015machine,sun2018reading} especially when there are only hundreds of documents available in a dataset such as MCTest~\cite{richardson2013mctest}. 
Along with the creation of large-scale reading comprehension datasets, recent machine reading comprehension models rely on end-to-end neural models, and it primarily uses word embeddings as input. However, ~\newcite{wang2016emergent,dhingra2017linguistic, dhingra2018neural} show that existing neural models do not fully take advantage of the linguistic knowledge, which is still valuable for machine reading comprehension. 
Besides widely used lexical features such as part-of-speech tags and named entity types~\cite{wang2016emergent,liu2017stochastic,dhingra2017linguistic, dhingra2018neural}, we consider more diverse types of external knowledge for performance improvements. Moreover, we accommodate external knowledge with probabilistic logic to potentially improve the interpretability of MRC models instead of using external knowledge as additional features.

\subsection{Explainable Machine Reading Comprehension and Question Answering}

To improve the interpretability of question answering, previous work utilize interpretable internal representations~\cite{palangi2017question} or reasoning networks that employ a hop-by-hop reasoning process dynamically~\cite{zhou2018interpretable}. A research line focuses on visualizing the whole derivation process from the natural language utterance to the final answer for question answering over knowledge bases~\cite{abujabal2017quint} or scientific word algebra problems~\cite{ling2017program}. ~\newcite{jansen2016s} extract explanations that describe the inference needed for elementary science questions (\eg, \emph{``What form of energy causes an ice cube to melt''}). In comparison, the derivation sequence is less apparent for open-domain questions, especially when they require external domain knowledge or multiple-sentence reasoning. To improve explainability, we can also check the attention map learned by neural readers~\cite{wang2016emergent}, however, attention map is learned in end-to-end fashion, which is different from our work.

A similar work proposed by~\newcite{sharp2017tell} also uses distant supervision to learn how to extract informative justifications. However, their experiments are primarily designed for factoid question answering, in which it is relatively easy to extract justifications since most questions are informative. In comparison, we focus on multi-choice MRC that requires deep understanding, and we pay particular attention to denoising strategies.
          
\section{Conclusions}

We focus on extracting evidence sentences for multiple-choice MRC tasks, which has not been studied before. We propose to apply distant supervision to noisy labels and apply a deep probabilistic logic framework that incorporates linguistic indicators for denoising noisy labels during training. To indirectly evaluate the quality of the extracted evidence sentences, we feed extracted evidence sentences as input to two existing neural readers. Experimental results show that we can achieve comparable or better performance on three multiple-choice MRC datasets, in comparison with the same readers taking as input the entire document. However, there still exist significant differences between the predicted sentences and ground truth sentences selected by humans, indicating the room for further improvements. 

\section*{Acknowledgments}
We thank the anonymous reviewers for their encouraging and helpful feedback.

\hhide{
\section*{Acknowledgments}
We thank Hoifung Poon for useful discussion on deep probabilistic logic. We thank Daniel Khashabi for help on evaluating our system on MultiRC test set. 
}


\bibliography{conll-2019}
\bibliographystyle{acl_natbib}

\clearpage
\newpage

\appendix
\section{Supplemental Material}
\label{sec:appendix}

\subsection{Deep Probabilistic Logic}

Since human-labeled evidence sentences are seldom available in existing machine reading comprehension datasets, we use distant supervision to generate weakly labeled evidence sentences: we know the correct answer options, then we can select the sentences in the reference document that have the highest information overlapping with the question and the correct answer option (Seciton~\ref{sec:transformer}). However, weakly labeled data generated by distant supervision is inevitably noisy~\cite{bing2015improving}, and therefore we need a denoising strategy that can leverage various sources of indirect supervision. 

In this paper, we use Deep Probabilistic Logic (DPL)~\cite{haidpl2018}, a unifying denoise framework that can efficiently model various indirect supervision by integrating probabilistic logic with deep learning. It consists of two modules: 1) a supervision module that represents indirect supervision using probabilistic logic; 2) a prediction module that uses deep neural networks to perform the downstream task. The label decisions derived from indirect supervision are modeled as latent variables and serve as the interface between the two modules. DPL combines three sources of indirect supervision: distant supervision, data programming, and joint inference. We introduce a set of labeling functions that are specified by simple rules, and each function assigns a label to an instance if the input satisfies certain conditions for data programming, and we introduce a set of high-order factors for joint inference. We will detail these sources of indirect supervision under our task setting in Section~\ref{sec:strategies}.


Formally, let $K=(\Phi_1,\cdots,\Phi_V)$ be a set of indirect supervision signals, which has been used to incorporate label preference and derived from prior knowledge. DPL comprises of a supervision module $\Phi$ over $K$ and a prediction module $\Psi$ over ($X$, $Y$), where $Y$ is latent in DPL:
\begin{eqnarray}
\small
P(K,Y \, | \, X)\propto \prod_{v}~\Phi_{v}(X, Y)\cdot\prod_i~\Psi(X_i, Y_i)
\end{eqnarray}

\eat{
\vspace{-8pt}
\[ L(\Phi, \Psi, X) = \math{E}_{y \sim \prod_v^{K}~\Phi_{v}(X, Y)} \ln \Psi(X_i, Y_i)\]
\vspace{-8pt}
}

Without loss of generality, we assume all indirect supervision are log-linear factors, which can be compactly represented by weighted first-order logical formulas \cite{richardson&domingos06}. Namely, $\Phi_v(X,Y)=\exp(w_v\cdot f_v(X,Y))$, where $f_v(X,Y)$ is a feature represented by a first-order logical formula, $w_{v}$ is a weight parameter for $f_v(X,Y)$ and is initialized according to our prior belief about how strong this feature is\footnote{Once initial weights can reasonably reflect our prior belief, the learning is stable.}. The optimization of DPL amounts to maximizing $\sum_{Y}P(K,Y|X)$ (\eg, variational EM formulation), and we can use EM-like learning approach to decompose the optimization over the supervision module and prediction module. See~\newcite{haidpl2018} for more details about optimization.

\eat{
At iteration $t$, we employ the following optimization procedure:
\begin{eqnarray}
\small
p^t(Y) \leftarrow \arg\min_{p}~D_{KL}(\prod_i~p_i(Y_i)~||\\ \nonumber
\prod_v~\Phi^{t-1}_v(X,Y)\cdot\prod_i~\Psi^{t-1}(X_i,Y_i)) \\  
\Phi^t \leftarrow \arg\min_{\Phi}~D_{KL}(p^t(Y)~||~ \prod_v~\Phi_v(X,Y))\\
\Psi^t \leftarrow  \arg\min_{\Psi}~D_{KL}(p^t(Y)~||~\prod_i~\Psi(X_i,Y_i))
\end{eqnarray}
}

\subsection{Denoising with DPL}

Besides distant supervision, DPL also includes data programming (\ie, $f_{v}(X, Y)$ in Section~\ref{sec:dpl}) and joint inference. As a preliminary attempt, we manually design a small number of sentence-level labeling functions for data programming and high-order factors for joint inference. 

For sentence-level functions, we consider lexical features (\ie, the sentence length, the entity types in a sentence, and sentence positions in a document), semantic features based on word and paraphrase embeddings and ConceptNet~\cite{speer2017conceptnet} triples, and rewards for each sentence from an existing neural reader, language inference model, and sentiment classifier, respectively.

For high-order factors, we consider factors including if whether adjacent sentences prefer the same label, the maximum distance between two evidence sentences that support the same question, and the token overlap between two evidence sentences that support different questions.

\begin{figure}[!h]
   \begin{center}
   \includegraphics[width=0.45\textwidth]{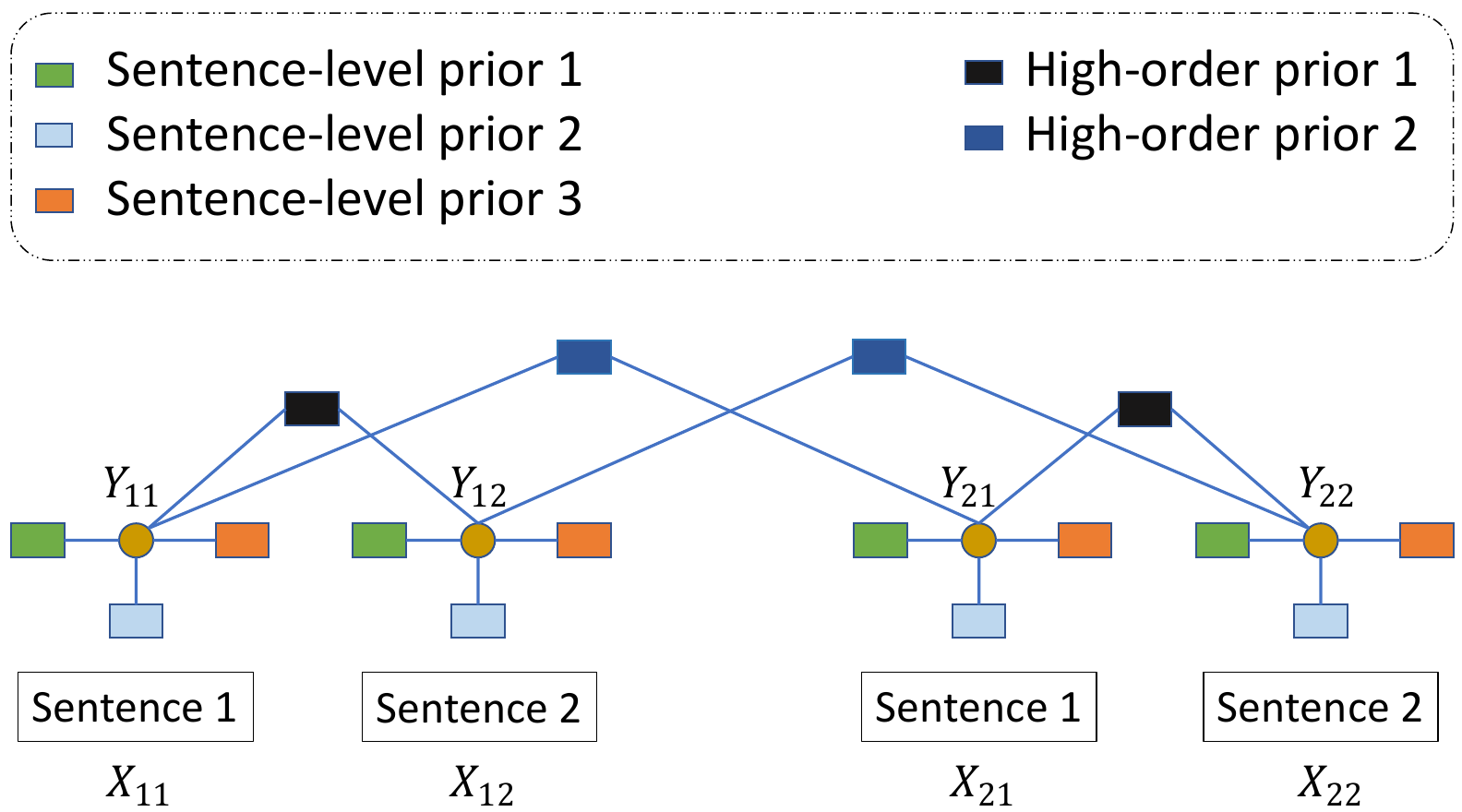}
   \end{center}
\caption{A simple factor graph for denoising.}
\label{fig:factorgraph}
\end{figure}

We show the factor graph for a toy example in Figure~\ref{fig:factorgraph}, where the document contains two sentences and two questions. $X_{ij}$ denotes an instance consisting of sentence $i$, question $j$ and its associated options, $Y_{ij}$ is a latent variable indicating the probability that sentence $i$ is an evidence sentence for question $j$. We build a factor graph for the document and all its associated questions jointly. By introducing the logic rules jointly over $X_{ij}$ and $Y_{ij}$, we can model the joint probability for $Y$.

\subsection{Indirect Supervision Strategies}
\label{sec:strategies}

Besides distant supervision, DPL also includes data programming and joint inference. For data programming, we design the following sentence-level labeling functions:

\subsubsection{Sentence-Level Labeling Functions}
\label{app:low}

\begin{itemize}
\setlength\itemsep{-0.25em}
\item Sentences contain the information asked in a question or not: for ``when"-questions, a sentence must contain at least one time expression; for ``who"-questions, a sentence must contain at least one person entity.  
\item Whether a sentence and the correct answer option have a similar length: $0.5 \leq \frac{\text{len(sentence)}}{\text{len(answer)}} \leq 3$.
\item A sentence that is neither too short nor too long since those sentences tend to be less informative or contain irrelevant information: $5 \leq \#~\text{of tokens in sentence} \leq 40$.
\item Reward for each sentence from a neural reader. We sample different sentences and use their probabilities of leading to the correct answer option as rewards. See Section~\ref{sec:implementation} for details about reward calculation.
\item Paraphrase embedding similarity between a question and each sentence in a document: $ \cos({e^{para}_{q}, e^{para}_{sent}}) \geq 0.4 $. 
\item Word embedding similarity between a question and each sentence in a document: $ \cos({e^{w}_{q}, e^{w}_{sent}}) \geq 0.3 $.
\item Whether question and sentence contain words that have the same entity type.
\item Whether a sentence and the question have the same sentiment classification result.
\item Language inference result between sentence and question: entail, contradiction, neutral.
\item \# of matched tokens between the concatenated question and candidate sentence with the triples in ConceptNet~\cite{speer2017conceptnet}: $\frac{\#~\text{of matching}}{\#~\text{of tokens in sentence}} \leq 0.2$.  
\item If a question requires the document-level understanding, we prefer the first or the last three sentences in the reference document.
\end{itemize}

\subsubsection{High-Order Factors}
\label{app:high}

For joint inference, we consider the following high-order factors $f_{v}(X, Y)$.
\begin{itemize}
\setlength\itemsep{-0.25em}
    \item Adjacent sentences tend to have the same label.
    \item Evidence sentences for the same question should be within window size $8$. For example, we assume $S_1$ and $S_{12}$ in Figure~\ref{fig:overview} are less likely to serve as evidence sentences for the same question.
    \item Overlap ratio between evidence sentences for different questions is smaller than $0.5$. We assume the same set of evidence sentences are less likely to support multiple questions.
\end{itemize}

\end{document}